%% file: roman-2020-mainprice.tex
\documentclass[letterpaper, 10 pt, conference]{ieeeconf}  

\IEEEoverridecommandlockouts                              

\usepackage{enumerate}
\usepackage{framed} 
\usepackage{todonotes}
\usepackage{balance}
\usepackage{subcaption}
\usepackage{multicol}
\usepackage{amsmath}
\usepackage{amssymb}
\usepackage{balance}
\usepackage{times} 
\usepackage{multirow}
\usepackage{tabularx}
\usepackage{color}
\usepackage{array}

\graphicspath{{figures/}}

\newcommand{\specialcell}[2][c]{%
  \begin{tabular}[#1]{@{}c@{}}#2\end{tabular}}
\newcolumntype{M}[1]{>{\centering\arraybackslash}m{#1}}

\graphicspath{{figures/}}
\input{basic_environments.tex}

\begin{document}

\title{\LARGE \bf
 An Interior Point Method Solving\\ Motion Planning
 Problems with Narrow Passages 
}

\author{Jim Mainprice$^{1, 2}$, Nathan Ratliff$^{4}$, Marc Toussaint$^{2, 3}$ and Stefan Schaal$^{5}$\\
\authorblockA{
	$^1$\tt{\small{firstname.lastname@ipvs.uni-stuttgart.de}}, 
	$^4$\tt{\small{nratliff@nvidia.com}},
	$^5$\tt{\small{schaal@google.com}}}
\authorblockA{$^1$Machine Learning and Robotics Lab, University of Stuttgart, Germany}
\authorblockA{$^2$Max Planck Institute for Intelligent Systems ;  IS-MPI ; T{\"u}bingen \& Stuttgart, Germany}
\authorblockA{$^3$Learning and Intelligent Systems Lab ;  TU Berlin ; Berlin, Germany}
\vspace{-.9cm}
}

\maketitle
\thispagestyle{empty}
\pagestyle{empty}

\begin{abstract}
Algorithmic solutions for the motion planning problem
have been investigated for five decades.
Since the development of A* in 1969
many approaches have been investigated,
traditionally classified as either grid decomposition,
potential fields or sampling-based.
In this work, we focus on using numerical optimization,
which is understudied for solving motion planning problems.
This lack of interest in the favor
of sampling-based methods is largely due to the
non-convexity introduced by narrow passages.
We address this shortcoming by grounding
the solution in differential geometry.
We demonstrate through a series of experiments
on 3 Dofs and 6 Dofs narrow passage problems,
how modeling explicitly
the underlying Riemannian manifold leads to
an efficient interior point non-linear programming solution.
\footnote{Jim Mainprice is the Interim Professor of the Machine Learning and
Robotics Laboratory with the University of Stuttgart. This work
was partially conducting as he was with Nathan Ratliff and Stefan Schaal at the Max Planck
Institute for Intelligent Systems, in T{\"u}bingen, Germany.}
\end{abstract}

\section{Introduction}

Autonomous motion planning is a central
component of autonomous behavior.
Hence, accuracy and effectiveness of motion planning
algorithms can have dramatic impacts in terms of safety
and acceptance of robots. Indeed, safety and more generally
human-robot interaction constraints are often modeled
as cost functionals \cite{Kruse:13}, which are in turn
optimized by a motion planning algorithm.

The motion planning community has focused
on sampling-based approaches over the last two
decades with a lot of success. The introduction
of probabilistically complete planners \cite{Lavalle:06} such as
Probabilistic Road Maps (PRMs) and Rapidly Exploring
Random Trees (RRTs) has allowed to solve virtually any problem
in any dimension. However, when applied to real world
robotics with many degrees of freedom in dynamic
environments, sampling-based algorithms are typically
too slow to be usable and many works usually
resort to local methods (i.e., potential fields \cite{Khatib:86}).

For manipulation, Motion Optimization (MO) \cite{Zucker:13}
is often preferred. In MO, motion planning
is solved using trajectory optimization, i.e., gradient-based
optimization in the full trajectory parameter space.
MO does not suffer the myopic shortcomings of local
methods as it considers the full horizon length,
while also using gradient information
provided by potential fields to
converge rapidly to a local minimum.
The disadvantage of MO is that it is remains local,
and performs poorly when the problem
is highly non-convex, where it often does
not converge at all.
Thus, MO is often used as a post-processioning step
 of an RRT search \cite{hauser:18, kuntz:20}.

The theory of motion planning is well developed and its connection
to differential geometry is long standing.
The geometry of the forward kinematics map is well
understood, and the issues linked to sampling or interpolating
SE(3) are well treated in classical text books \cite{Latombe:92, Lavalle:06}.
However due to the large interest in sampling-based approaches
in the last two decades, little effort has been made to
understand the differential structure of the workspace geometry,
which, we argue, is essential for motion optimization. 
One notable exception is the work of Ratliff et al.
\cite{ratliff2015understanding}, where the notion
of Natural attractors and Riemannian metrics
that we further improve in this work have been initially introduced.

\begin{figure}
\centering
\includegraphics[width=1\linewidth]{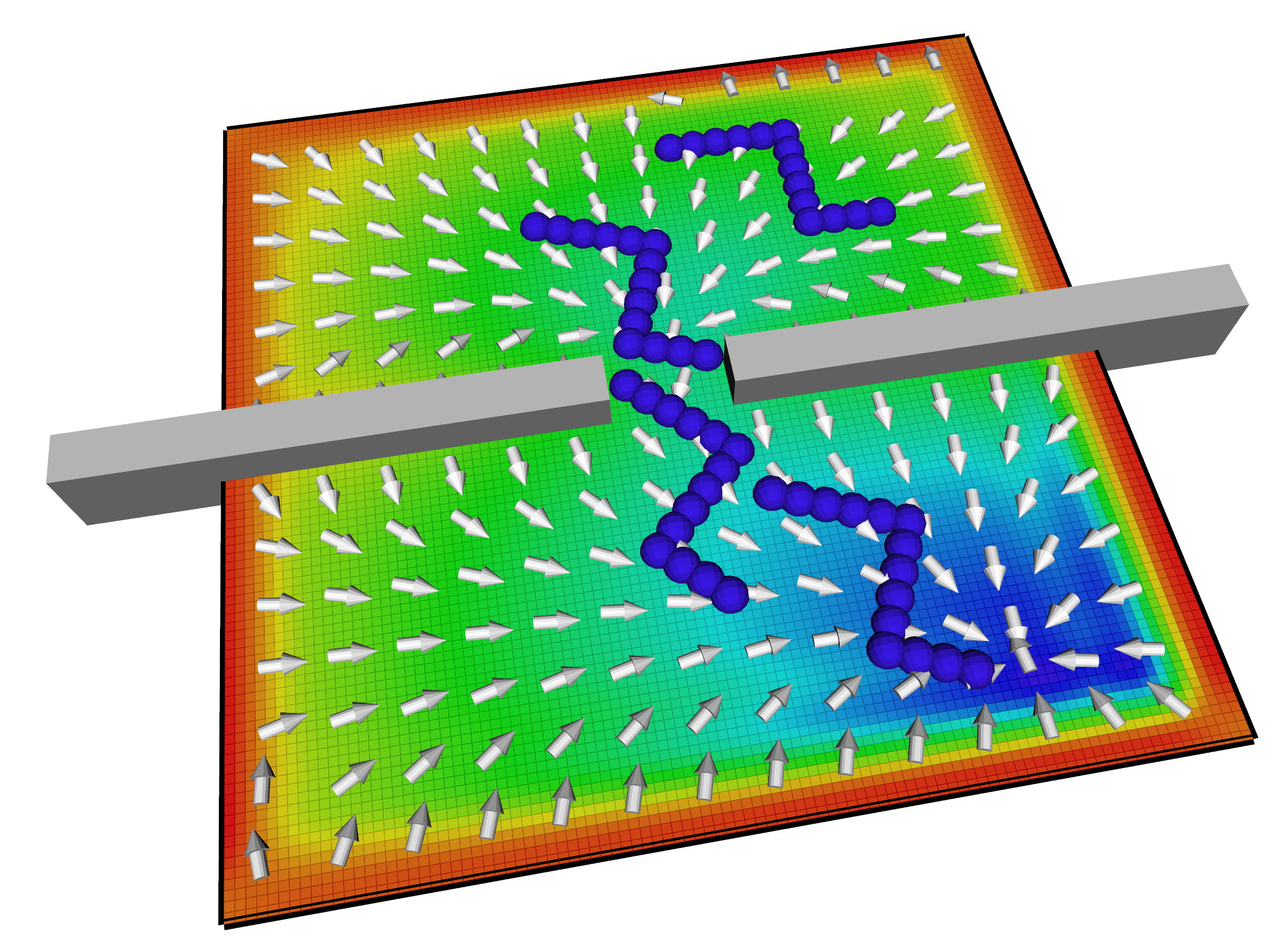}
\caption{Geodesic flow attractor in a planar environment, cool and hot colors are small and large geodesic distances resp.}
\label{fig:geodesic_flow}
\vspace{-.5cm}
\end{figure}

In this work we introduce a formal treatment of the workspace
geometry in terms of Riemannian geometry, and propose
new terms to model the goal constraint, i.e., geodesic attractor
(see Figure \ref{fig:geodesic_flow}),
and a geodesic flow agreement terms.
We show experimentally
that these terms lead to better convergence rates on three narrow passage environments, involving 3 and 6 Dofs.
To our knowledge this work is the first to present the
implementation of an Interior Point algorithm to
solve complex path planning problems
usually treated by sampling-based methods.

The remainder of this paper is structured as follows: In the next section we give a description of related work.
In Section \ref{sec:riemannian_metric}, we formalize
and discuss the notion of Workspace Riemannian Metrics. In Section \ref{sec:geodesic_flow},
we describe the geodesic flow which is then used
to introduce a new constraint model and an objective term.
In Section \ref{sec:results}, we present empirical results on three narrow passage problems which demonstrate
the efficacy of our approach.

\section{Related Work}
\label{sec:related_work}

\subsection{Motion optimization}


Motion optimization \cite{Zucker:13, Kalakrishnan:11, Schulman:13, Toussaint:14} relates to functional optimization algorithms,
which take gradient space in trajectory space. This approach was originally pioneered by Quinlan and Khatib \cite{Quinlan:93}, and further by Brock \cite{Brock:00}, with the aim to produce reactive robotic behaviors. These early works modeled the trajectory by a mass-spring system, and motion planning was solved by simulating the virtual elastic band system. In contrast to this initial approach, motion optimization models the problem through numerical optimization, leveraging the large body of work
in the area.

In \cite{Ratliff:15}, Ratliff and Toussaint, have proposed
to minimize the geodesic distance of the body part
instead of the Cartesian space arc length
proposed in \cite{Zucker:13}.
These details are crucial for achieving
convergence of the non-linear program for motion planning.
In \cite{Mainprice:16}, we have proposed a Riemannian
metric for handling arbitrary workspace geometries that
go beyond primitive shapes such as circles and boxes
by defining harmonic potentials.
In this work we go a step further and address the problem
of geodesic distance to the goal, by leveraging the heat
method \cite{Crane:13}.


\subsection{Harmonic functions}

Originally artificial potential fields were used for the obstacle avoidance function within the robot workspace \cite{Khatib:86}. This allowed to get realtime robot behavior by using the operational space formalism. Potential fields are known to be prone to local minima.
Thus functions that limit local minima by satisfying
Laplace's Equation have been investigated \cite{Connolly:93} (i.e., harmonic functions).

Laplace's equation is a partial differential equation, so numerically solving for a harmonic potential field can be computationally challenging and suffers from the curse of dimensionality when applied to the robot configuration space. 
Recently \cite{Wray:16}, this method has been extend
for computing robot navigation paths by leveraging GPUs.
In this work we combine such functions
with motion-optimization techniques.
For this we make use of the heat method \cite{Crane:13}, which diffuses heat in the workspace leading to harmonic function at convergence.

\subsection{Numerical Optimization}

The motion objective is generally optimized using an augmented Lagrangian formulation,
which uses generic constrained optimization solvers, recomputing the Lagrange multipliers in the outer loops and constructing a series of unconstrained objectives for the inner loop optimizers. These unconstrained objectives encode the violated constraints as shifted penalties.
In this work we instead use an Primal-dual interior-point method \cite{Wachter:06} for nonlinear optimization based on a Gauss-Newton approximate of the
Hessian. This algorithm handles better inequality constraints by using
log barrier terms and forcing the solution to remain within
the feasibility region.

\section{Workspace Riemannian Metrics}
\label{sec:riemannian_metric}
The whole problem of path planning is to take into account the obstacles 
$\spt{O}_i$ that populate the workspace  $\spt{W} \subset \mathbb{R}^3$. 
Obstacles regions are subsets of $\spt{W}$. Obstacles are define to have
non empty interior and have a smooth boundary $\partial \spt{O}_i$.

The freespace is the set difference 
of $\spt{W}$ and the union of the obstacles:
$$
\spt{F} := \spt{W}_{free} = \spt{W} \setminus \bigcup_{i \in \spt{O}} \spt{O}_i.
$$
The freespace is a smooth compact manifold with a smooth boundary $\partial \spt{F}$. It is associated with an atlas of a single chart, which is the global Euclidean coordinate system. 

A metric tensor $g_p$ is a smooth map defined on the tangent bundle of a manifold $\spt{M}$. The map $g_p$ is defined for each point $p \in \spt{M}$, and associates a real number given two vectors $g(p) : T_p(\spt{M}) \times T_p(\spt{M}) \rightarrow \mathbb{R}$. A smooth manifold equipped with a positive definite metric tensor, i.e., $\forall (u, v) \in T_p(\spt{M})^2, g_p(u,v) > 0$, is called a Riemannian manifold.

For points $ p \in \spt{W}$, the tangent space $T_p(\spt{W})$ is the Euclidean space.
If $\spt{F}$ is equipped with the usual Euclidean metric $\| . \|_2$,
it defines a Riemannian manifold $(\spt{F},g = I)$. 
Here, we aim to characterize Riemannian metrics tensors $g$ for $\spt{F}$, which make geodesics wrap around obstacles regions $\spt{O} =\bigcup_{i \in \spt{O}} \spt{O}_i$.
The following sections will identify a class of functions that induces such metrics.

\subsection{Define $g$ over $\spt{W}$ or $\spt{F}$ ?}

The obstacle region $\spt{O}$  defines disconnected subsets of $\spt{W}$, which can be viewed as topological holes. In other words, the freespace is itself a Riemannian manifold $\spt{F}_{rie}$ in which all geodesics naturally avoid $\spt{O}$. 
Recall that a geodesic on a Riemannian manifold $\spt{M}$ is defined to be the a curve 
$\gamma  : \mathbb{R}^+ \rightarrow \spt{M}$, which ``length" is measured as:

$$
L(\gamma) = \int^{T}_{0} \sqrt{g_{\gamma(t)} (\dot \gamma(t), \dot \gamma(t))} dt
$$

\noindent
where $\forall t , \gamma(t) \in \spt{M}$ and is continuously differentiable. 
Geodesics correspond to curves for which $\nabla_{\dot{\gamma}} \dot{\gamma} = 0$, where $\nabla$ is an affine connection (intuitively this means that the acceleration is either 0 or orthogonal to the tangent plane, it generalizes the notion of straight line to curved spaces). Geodesics are invariant to affine re-parametrization ($t' = at + b$), but not to arbitrary re-parametrizations. 

There may exist multiple geodesics linking two points in space $p_1, p_2 \in \spt{M}$. Minimizing geodesics define a metric over Riemannian manifolds: $d(p_1, p_2) = \text{argmin} L(\gamma)$. These curves can be obtained by the calculus of variation on the energy functional because:
$$
L(\gamma) \leq 2(p_1 - p_2) E(\gamma).
$$

Hence by definition, minimizing geodesic are curves that only take value in the manifold. This implies that we can consider two types of problems to have minimizing geodesic wrap around obstacles:

\begin{framed}
\begin{enumerate}[a)]
\setlength\itemsep{.01cm}
\item \textit{Topological}: $g$ defined over $\spt{W}$, such that 
$\forall p_1, p_2 \in \spt{F} \implies \forall t \in [p_1, p_2], \gamma(t) \in \spt{F}$
\item \textit{Behavioral}: $g$ defined over $\spt{F}$, with $\gamma(t) \in \spt{F}$ by definition
\end{enumerate}
\end{framed}

Finding a satisfying definition for each of these conditions is different. For a) the problem is to fine the constraints in $g$ such that the implication is true. For b) we need to define what a good metric $g$ would be.

%

\subsection{The Obstacle-based Riemannian metric}

To define a metric $g$ in the line of a):
\begin{quote}
`` \textit{$g$ defined over $\spt{W}$, such that 
$\forall (p_1, p_2) \in \spt{F}^2 \implies \forall t \in [p_1, p_2], \gamma(t) \in \spt{F}$} "
\end{quote}
poses constraints on the geodesic flow inside $\spt{F}$ induced by the metric. Generally,
this is enforced by having $\forall (p_1, p_2) \in \spt{O}^2,\lim_{p_1 \to p_2} L(\gamma)  > \text{diam}(\spt{F})$. This will ensure that any minimizing geodesic in $\spt{F}$ is shorter than the minimizing geodesics in $\spt{O}$.

We define the following properties of the geodesics in a Riemannian manifold
equiped with such a metric:

\begin{framed}
\begin{enumerate}[i)]
\setlength\itemsep{.01cm}
\item \textit{Non-penetrating:} At the boundary, the geodesics should be parallel to the boundary. Thus if $\hat n$ is normal to $p \in \partial \spt{F}$ pointing inward and $\spt{V} = \{ v | \hat n^T v > 0 \}$ is the set of all vectors penetrating the $\spt{O}$, then $\forall v \in \spt{V}, g_p(v, v) = 0$. In other words, the null space of $g_p$ should contain all vectors with positive dot product with boundary normal vectors.
\item \textit{Blending to Euclidean:} Away from the boundary geodesics should obey Euclidean geometry. $\lim_{d(p) \to \infty} g_p = \mathbb{I}$, where $d(p)$ is the minimal distance to the boundary at $p$. 
\item \textit{Multi-resolution:} The influence of the detail geometry of the boundary should vanish with distance from the boundary.
\item \textit{Ordering:} On a planar section of the workspace, a geodesic between two points farther from the boundary than points closer should not cross. 
\end{enumerate}
\end{framed}

%


\subsection{The Eigen spectrum of the metric tensor}

In order to gain intutition on what the obstacle-based metric tensor represent
we can look at the behavior of its eigen spectrum.
The tensor stretches space along particular dimensions.
A Riemannian metric tensor $g$ operates on the tangent space $T_p(\spt{M})$
(i.e., the space of velocities) of a given manifold point $p \in \spt{M}$.

If the manifold $\spt{M} \subset \mathbb{R}^3$, and the tensor $g$ is a scaling of the dot product:
$$
g_p(u, v) = u^T A_p v 
$$
\noindent
where $v,w \in T_p(\spt{M})$.
Since the matrix $A$ is semi-positive definite by definition,
the singular value decomposition has the following form:
$$
A_p = U \Sigma U^T = 
U 
\begin{pmatrix} 
\lambda_1 & 0  & 0 \\ 
0 & \lambda_2  & 0 \\ 
0 & 0  & \lambda_3 \\  \end{pmatrix}
U^T
$$
where $\lambda_i$ are the eigenvalues of $A$. The matrix $U$ is orthogonal, it operates a change of coordinates that preserves scaling. The matrix $\Sigma$ scales each dimension in that coordinate system. Thus the eigen values describe how the space is warped locally to $p$.

In the case where $\spt{M}$ is a subset of Euclidean space,
the tangent space is $T_p(\spt{M}) = \mathbb{R}^3$.
The metric tensor operates a scaling of euclidean space,
define by the eigen spectrum of $g_p$. 
At the boundary $\partial \spt{F}$ the
inverse of the metric tensor would
loose rank, effectively removing volume in the tangent
space going through the obstacle surface.
Equivalently, $g_p$ measures velocity in
the direction of the surface infinitly.

Now that we have defined what a good
Workspace Riemannian metric might be
and understand how they operate in the workspace,
we can introduce two key elements that
we use in our experiments.
First the workspace geometry map $\phi_{\scalebox{.5}{WS}}(p) $, which allows to define a metric tensor $g$ and
the geodesic flow, which encodes the geodesic
distance to the goal.

\subsection{The workspace geometry map}

Any mapping of the form $z = \phi(p)$
into Euclidean space defines a Pullback metric
$A(p) = J_{\phi}^T J_{\phi}$. 
In fact, we can represent any metric $A(p)$,
as the pullback of a mapping $\phi(p)$ to some higher dimensional space \cite{Ratliff:15}. This is 
the famous Nash embedding theorem.
This observation means that the generalized velocity term of the form $\dot p^T A(p) \dot p$ can be equally well described as a Euclidean velocity through the map's co-domain. 

The metric $A(p)$ can be used to generate terminal potentials that follow geodesic contours of the workspace under the metric.
We denote this as the Natural attractor,
as the gradient of this attractor is the Natural gradient.
Using a map $\phi(p)$ allows a more convenient representation that is easier to use to form attractors for the terminal potential than directly specifying $A(p)$.

In \cite{Mainprice:16},
we make use of the electric potential proposed in \cite{Wang:13}.
The potential field emanating from an object surface $\Omega$
can be used to define a coordinate system around the object.
Three values are used to define a coordinates, i.e.,
the potential and two coordinates for the field lines.

This coordinate system would provide a good candidate for the map $\phi$,
however the metric induced by such a map contains a seam
(i.e., a point where the $u$ and $v$ coordinates wrap from 0 to 2$\pi$).
Additionally, potential-based metrics induce lower dimensional $\phi$
maps which are less expensive to compute. Thus our workspace geometry map is of the form:

\begin{equation}
\phi_{\scalebox{.5}{WS}}(p) = 
\left[
\begin{array}{c}
\alpha_1 \phi_1(p)\\
\vdots \\
\alpha_{d+1}\phi_{d+1}(p)\\
\end{array}
\right],
\label{eq:workspace_map}
\end{equation}

\noindent
where $\phi_i$ are the individual potential values computed for each object in the scene. $\phi_{d+1}$ is the 3D identity map and $\alpha_{i}$ are proximity functions, which is constant for $i = d+1$.
In our experiments, we simply use
$\phi_i \sim \exp\big(-\sigma(p) \big)$,
where
$\sigma : \mathbb{R}^N \to \mathbb{R}$, is a signed
distance function, negative inside obstacles and positive outside.

\section{Geodesic flow}
\label{sec:geodesic_flow}
At a particular point $p$
on the manifold, the geodesic flow is defined as
$$G^t (V) = \dot{\gamma}_V (t)$$
\noindent
where $V \in T_p(\spt{W})$ is a vector on the tangent space.
Hence it defines all the points on the manifold that can
be reached in length $t$.
Computing the geodesic
flow seems prohibitive as they
would generally require either shooting geodesics
from the source point $p$ or solving many shortest
path problems.

However geodesic flows can be helpful in solving
motion optimization problems as they provide real
geodesic distance, which would wrap correctly around
obstacles, informing the optimizer about the true
nature of the underlying manifold.

Hence in this work, we compute this flow using the heat method \cite{Crane:13},
where heat is diffused on a regular grid according
to the heat equation:
$$\frac{\partial \phi}{\partial t} = \Delta \phi = \sum^N_{i=1} \frac{\partial \phi}{\partial x_N}$$
\noindent
where $\Delta$ is the Laplace operator and $\phi$ is some potential function, i.e., heat. It is possible to either solve
the heat equation in closed form,
resulting in a matrix inversion operation, or to iterate over the grid,
similar to policy evaluation in dynamic programming.
In either case only small incremental steps can be taken in time.

The gradient of the resulting potential function
agrees with the geodesic distance gradient,
which stems from the property of diffusion processes.
Note that at convergence $\Delta \phi = 0$, which means
that $\phi$ is a harmonic function, comprising a single
maximum, i.e., the heat source.

\begin{figure}[b]
\centering
\includegraphics[width=1\linewidth]{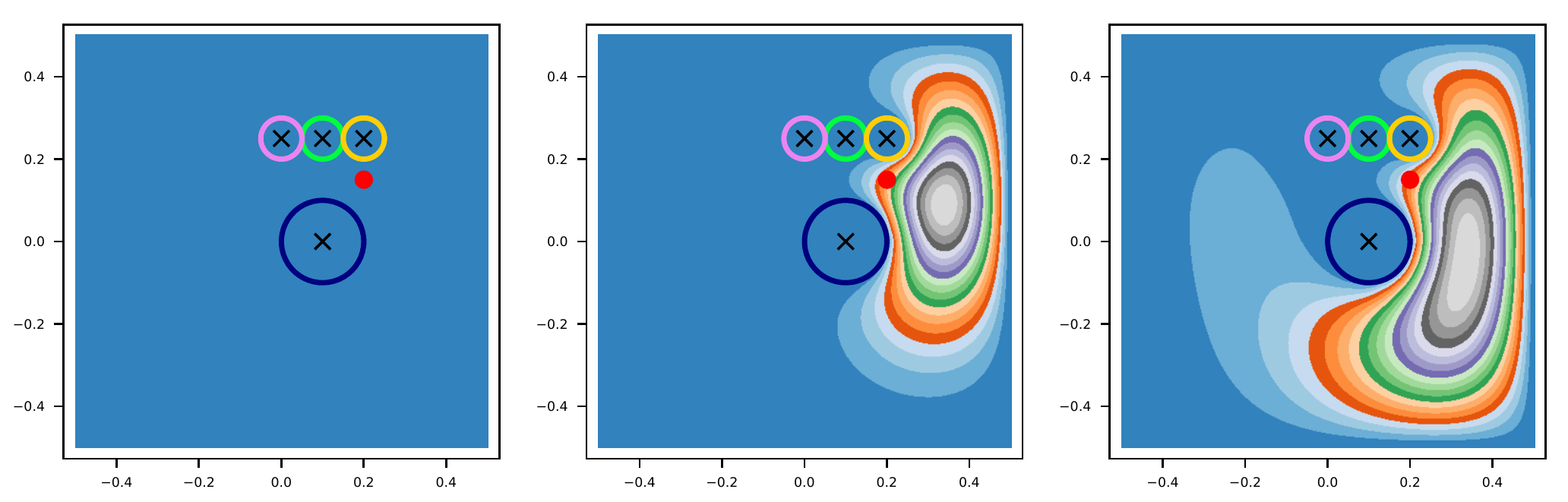}
\caption{Heat diffusion process in an environment with circular obstacle. The heat source is initially set to the red dot (left) and flows through the environment by following the heat equation.}
\label{fig:heat}
\vspace{-.4cm}
\end{figure}

Figure \ref{fig:heat} shows a heat diffusion process
calculated in a two dimension workspace populated by three
circular obstacles. Here the heat source is only set at the first
time step and then left out. When computing a geodesic flow
the heat source is kept at a fixed temperature.

In order to retrieve a geodesic distance we
solve for the potential by inverting a matrix relating the potential to its spacial derivative.
The geodesic attractor is then obtained we invert the resulting vector field to point
towards the source and interpolating that vector field on a
regular grid using bi-cubic and tri-cubic
splines in the planar and Cartesian case
respectively. The resulting vector field
can be seen for the planar case in Figure \ref{fig:geodesic_flow}.

Note that we performed diffusion on low resolution grid
in all our experiments in a time negligible
compared to the optimization time.
The sink of the attractor is often inaccurate leading to
bad convergence of the interior point algorithm.
Hence we linearly blend the geodesic attractor
to a euclidean attractor close to the goal.

\section{Results}
\label{sec:results}

Our experiments are conducted to highlight the
impact of considering explicitly
the Riemannian manifold structure of the workspace
in motion optimization.

\subsection{Setup}

We first defined three environments
that exhibit narrow passage to a certain degree.
We planed motions using freeflying robots with
3 Dofs in the planar case and 6 Dofs in the Cartesian case.
The forward kinematics maps $x : q \mapsto p \in \mathbb{R}^N$ are defined
using a single homogeneous transform, for which we define
analytically the Jacobian. In the Cartesian case, i.e., 3D,
we parametrize the robot orientation with Euler's angles
and use a rotation matrix internally.
Each robot's body is represented by $\approx$ 10 keypoints
with radius, as can be seen in Figure \ref{fig:environments}.

In Table \ref{tab:objective_terms}
we report the different objective terms used in our experiments.
The acceleration terms favors shorter and smoother
paths. The joint postural serves to resolve the redundancy
at the goal configuration.
Additionally to these objectives, we define two types
geodesic terms, one based on the workspace
geometry map $\phi_{\scalebox{.5}{WS}}$,
which we used
in our previous work \cite{Mainprice:16}, to implicitly
define the metric tensor.
In this work we additionally introduce the 
objective term ``geodesic flow" which measures
the agreement between the motion of the keypoints
$\dot x_t^i$ and the geodesic flow.
The equation presented here is a simplified version
of the objective term which uses a linearization
of arccos.

In Table \ref{tab:constraint_terms} we introduce
the two types of constraint functions that we
make use of for goal and obstacle avoidance.
The signed distance functions $\sigma$ are used
to define collision constraint.
All our obstacles are boxes represented by
an exact box distance, for which we define gradient
and hessian. We bound the hessian near
the edges of the box for numerical stability.
We also define a softmin
function which allows to define one collision constraint per configuration
rather than one constraint per keypoint and configuration.
We found this to make the algorithm more stable in practice.
The temperature parameter of the softmin is set very aggressive to approximate
a min function very closely.

\begin{table}[t]
\begin{center}
\begin{tabular}{|c|c|}
\hline
 & \\[-0.25cm]
Objective Terms & Mathematical Expressions \\
\hline
 & \\[-0.25cm]
\specialcell{Squared-norm of\\ C-space accelerations} 
&  $c_{t1}(q_t, \dot q_t, \ddot q_t) = \| \ddot q_t \|^2$  \\
\hline
 & \\[-0.2cm]
\specialcell{Geodesic Term} 
& $c_{t2}(q_t, \dot q_t, \ddot q_t) = \| \frac{d}{dt} \phi_{\scalebox{.5}{WS}}\big( x(q_t) \big) \|^2$ \\[0.1cm]
 \hline
 & \\[-0.2cm]
\specialcell{Geodesic Flow Term} 
& $c_{t3}(q_t, \dot q_t, \ddot q_t) = 
\arccos( \dot x_t^T \frac{\partial \phi_{\scalebox{.5}{F}}\big( x(q_t) \big) }{\partial x})$ \\[0.1cm]
\hline
 & \\[-0.1cm]
Joint postural & 
$c_{t4}(q_t, \dot q_t, \ddot q_t) = \| q_t - q_{\text{default}} \|^2$ \\[0.1cm]
\hline
\end{tabular}
\end{center}
\caption{Elementary objective terms}
\label{tab:objective_terms}
\end{table} 

\begin{table}[t]
\begin{center}
\begin{tabular}{|c|c|}
 \hline
Constraint Terms & Mathematical Expressions \\
\hline
 & \\[-0.25cm]
\specialcell{Signed distance function\\ in workspace geometry} 
& $c_{t5}(q_t)  = \text{softmin}_i \sigma[ \big( x(q_t) \big) ]$ \\
\hline
 & \\[-0.25cm]
\specialcell{Euclidean goal} 
& $c_{t6}(q_T)  = \| x(q_T) - x_{\text{goal}} \|$ \\
\specialcell{Natural goal} 
& $c_{t7}(q_T)  = \| 
\phi_{\scalebox{.5}{WS}}(\big( x(q_T) \big)) - 
\phi_{\scalebox{.5}{WS}}\big(x_{\text{goal}} \big) \|$ \\
\specialcell{Geodesic goal}
& $c_{t8}(q_T)  =\phi_{\scalebox{.5}{F}}(\big( x(q_T) \big))$ \\
\hline
\end{tabular}
\end{center}
\caption{Elementary constraint terms}
\vspace{-0.5cm}
\label{tab:constraint_terms}
\end{table} 

Finally we use three different goal equality constraints
for comparison.
The goal constraints are all defined with respect
to a single keypoint serving as end-effector of the free-flying
robot. We define a vanilla Euclidean attractor that simply
computes euclidean distance to the goal. We then define
a Natural attractor that defines the euclidean distance
in the workspace map $\phi_{\scalebox{.5}{WS}}$.
Finally we define our geodesic distance attractor $\phi_F$,
as depicted Figure \ref{fig:geodesic_flow}.

We make use of  the interior point algorithm
IPOPT \cite{Wachter:06} for optimizing the objective functional.
The functional is defined similarly to the KOMO objective
introduced in \cite{Toussaint:14}.
We define a clique at each time step,
for which we compute velocity and accelerations 
by finite differences.

\subsection{Statistical study}

\begin{figure*}
\centering
\includegraphics[width=.23\linewidth]{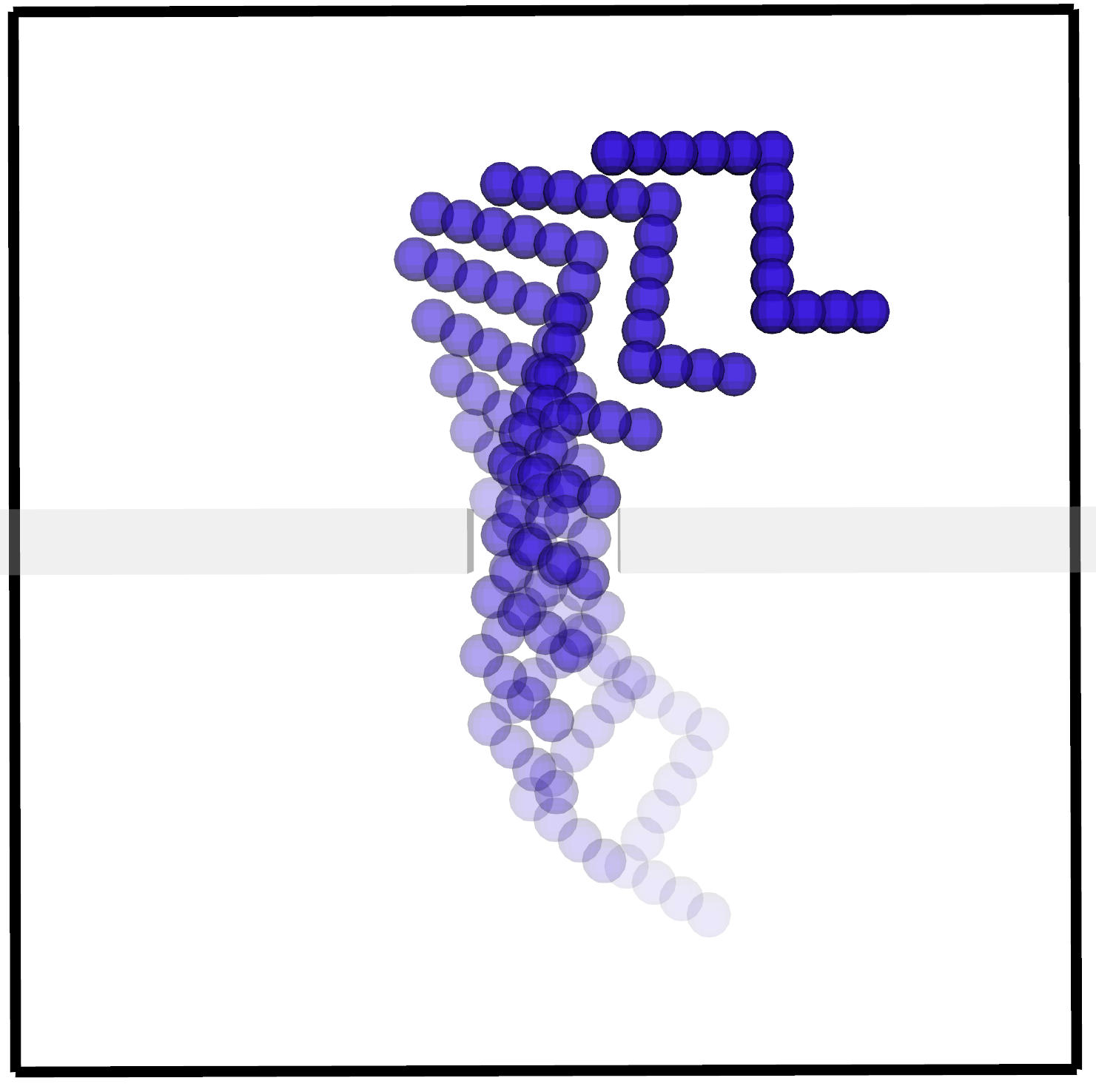}
\includegraphics[width=.23\linewidth]{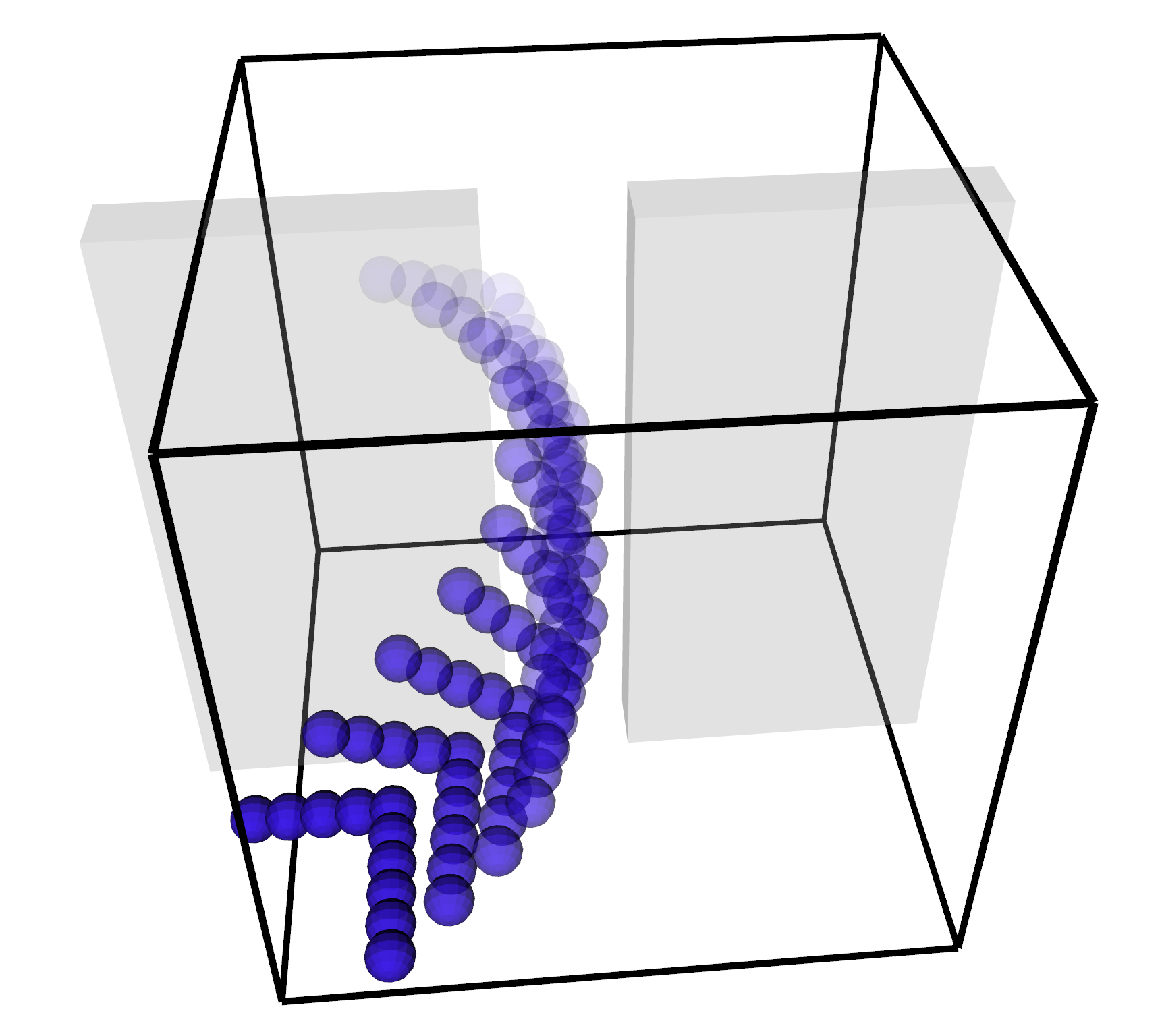}
\includegraphics[width=.23\linewidth]{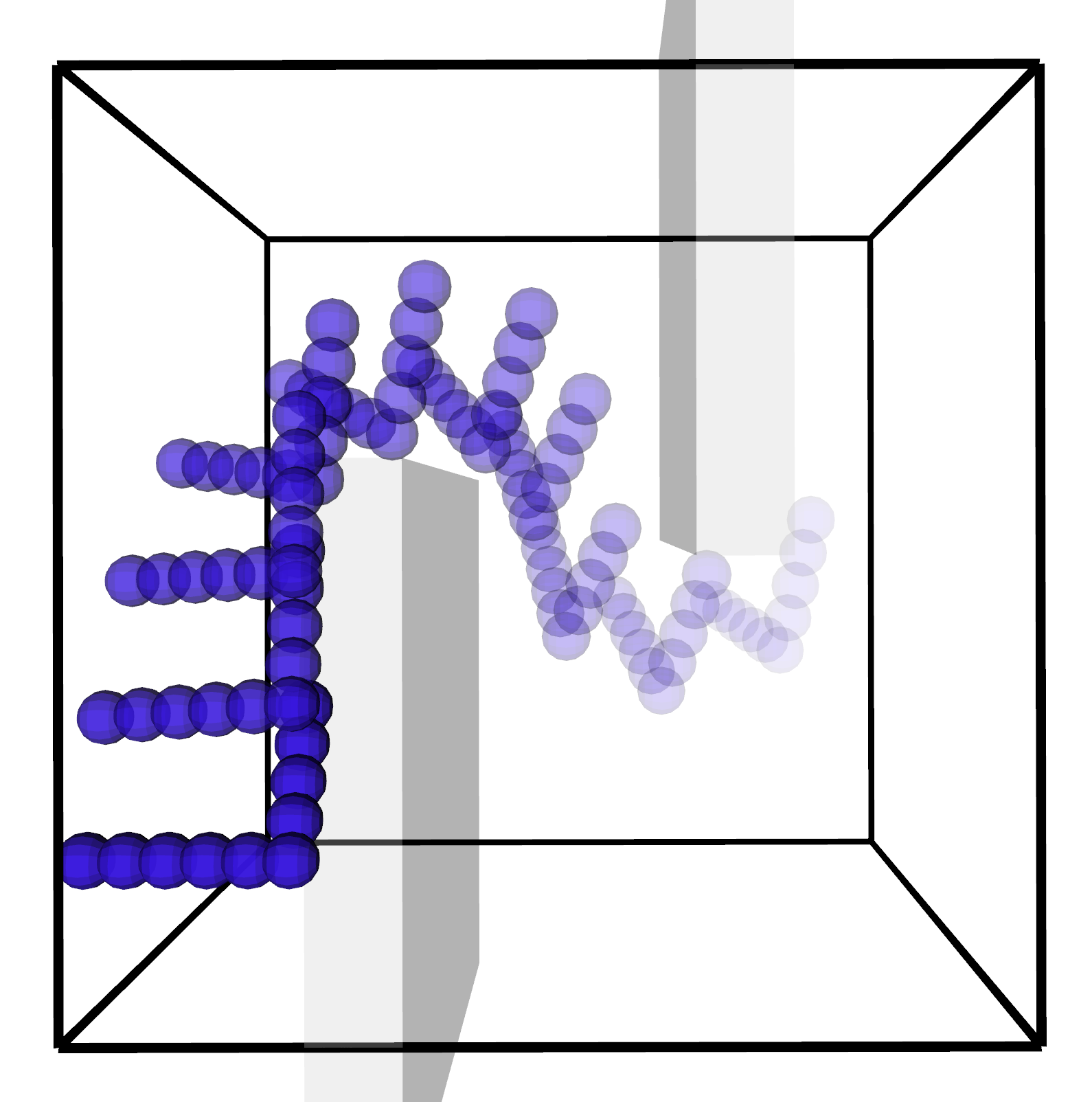}
\includegraphics[width=.23\linewidth]{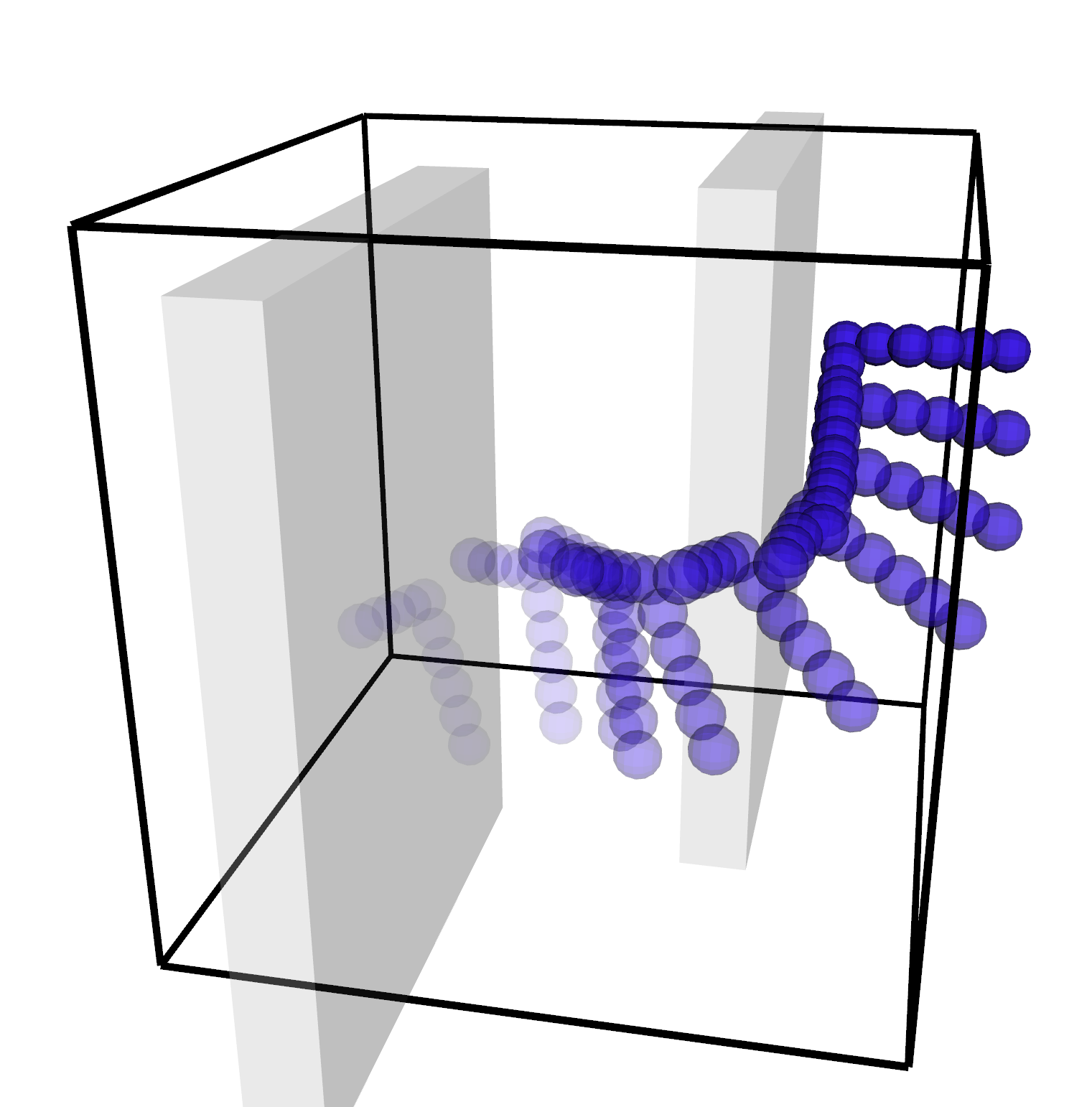}
\caption{2D planar environment with SE(2) configuration space, translation and rotation (left), and two 3D Cartesian environments (right) with free-flying SE(3) configuration spaces used in the experiments (one view angle of Cartesian Narrow and two view angles Cartesian Maze) .
The trajectories show time frames as color fading.
All environments present a narrow passage, which requires to
find a motion coordinating translation and orientation DoFs to find a collision free motion. }
\label{fig:environments}
\vspace{-.7cm}
\end{figure*}

In order to assess the influence of the different
terms on the ability of the optimizer to find
feasible motion plans we conducted an ablation
study. In each case we start from a single
configuration which is depicted in strong
blue in the motion traces of Figure \ref{fig:environments}.
We then sampled a goalset on the other side of the narrow passage
on a regular grid.

\begin{table}[h!]
\begin{center}
\caption{Goals sampled per environments}
\begin{tabular}{c|c}
Planar Narrow & 28 \\
Cartesian Narrow & 32 \\
Cartesian Maze & 36 \\
\end{tabular}
\end{center}
\vspace{-.5cm}
\end{table}
Each environment is tested with 9 conditions
except for the planar case where we also report
using the geodesic flow objective.
This results in about 250 trajectories per environment,
which took under one hour distributed over three cores.
All our implementation relies on c++, but it is
not optimized for efficiency, we do not report times
as our focus is to asses success rates among
different modeling paradigms. We stopped
the optimizer after 20 seconds if not converged.
Success is defined as reaching the goal and being
collision free.
We also report the break down in Table \ref{tab:results}.
The best success rate for each setup is highlighted
in bold.

\begin{table*}
\begin{center}
\scriptsize
\caption{Simulation results on the three problems of Figure \ref{fig:environments}.
All numbers are rates averaged over a goal set. }
\label{tab:results}
\begin{tabular}{|c|c|c|c|c|c|c|c|c|c|c|c|c|c|c|c|c|c|c|}
\hline
Attractor & 
\multicolumn{3}{|c|}{Euclidean} & \multicolumn{3}{c|}{Natural}  & \multicolumn{3}{c|}{Geodesic Flow} &
\multicolumn{3}{|c|}{Euclidean} &  \multicolumn{3}{c|}{Natural}  & \multicolumn{3}{c|}{Geodesic Flow} \\
Geod. Term & 0 & 10 & 50 & 0 & 10 & 50 & 0 & 10 & 50 &
			 0 & 10 & 50 & 0 & 10 & 50 & 0 & 10 & 50 \\
\hline
\hline
Problem & \multicolumn{9}{|c|}{Planar Narrow} & \multicolumn{9}{|c|}{Cartesian Narrow} \\
\hline
success & 0.00 & 0.14 & 0.04 & 0.07 & 0.00 & 0.00 & 0.04 & 0.18 & \textbf{0.29}  &
   0.00 & 0.62 & 0.62 & 0.00 & 0.06 & 0.28 & 0.00 & 0.88 & \textbf{0.91} \\
collision free & 0.00 & 0.14 & 0.04 & 0.57 & 0.57 & 0.71 & 0.04 & 0.25 & 0.54 &
   0.03 & 0.72 & 0.69 & 0.69 & 0.81 & 0.94 & 0.00 & 0.88 & 0.97 \\
goal reached & 1.00 & 1.00 & 1.00 & 0.39 & 0.36 & 0.21 & 0.96 & 0.89 & 0.71 &
   0.97 & 0.88 & 0.94 & 0.31 & 0.25 & 0.34 & 1.00 & 1.00 & 0.94 \\
\hline
\hline
Problem & \multicolumn{9}{|c|}{Planar Narrow (with flow)} & \multicolumn{9}{|c|}{Cartesian Maze} \\
\hline
success & 0.00 & 0.18 & \textbf{0.32} & 0.00 & 0.04 & 0.14 & 0.00 & 0.14 & \textbf{0.32} &
  0.00 & 0.06 & 0.42 & 0.00 & 0.03 & 0.19 & 0.00 & 0.31 & \textbf{0.53} \\
collision free & 0.04 & 0.29 & 0.46 & 0.18 & 0.18 & 0.54 & 0.00 & 0.14 & 0.32 &
  0.00 & 0.08 & 0.50 & 0.56 & 0.64 & 0.83 & 0.00 & 0.31 & 0.69 \\
goal reached & 0.93 & 0.86 & 0.82 & 0.68 & 0.57 & 0.43 & 0.75 & 0.86 & 0.96 &
  0.97 & 0.88 & 0.94 & 0.31 & 0.25 & 0.34 & 1.00 & 1.00 & 0.94 \\
\hline
%

%

%

\end{tabular}
\end{center}
\vspace{-.3cm}
\end{table*}

\subsection{Comparing different goal constraints}

We first compare the different goal constraints.
The geodesic flow attractor which we introduce in this paper
outperforms Euclidean and Natural on the three benchmarks.

The Natural attractor performs the worst.
The gradient of the attractor vanishes
due to the flat geometry of the boxes obstacles.
In environments populated
with more round obstacles Natural attractors
can perform significantly better than Euclidean attractors
pulling the motion towards the goal by wrapping around obstacles.
This problem is highlighted by the \textit{goal reached}
rate which are low with all Natural attractors.
On the other hand the geodesic attractor is able to pull
correctly the freeflyers into the passage leading
to \textit{goal reached} rates of near 100\% in both Cartesian case.

\subsection{Comparing with and without flow}

A narrow passage forces the solution motion to coordinate
the different DoFs of the robot to navigate the passage.
We test a heuristic forcing the motion
to follow the geodesic flow towards the narrow passage
on the planar case, which presents a more challenging
case as one can see by the low success rates in this case.

Forcing the motion to agree with the geodesic
flow enhances very significantly the case where
the goal constraint does not allow to find good passages
through the narrow passage, which is especially the
case with the Euclidean attractor.

\subsection{Riemannian metrics}

We also study the geodesic term
based on the workspace geometry map.
This term was introduced
in \cite{Ratliff:15}, but no formal experiments
were conducted to measure the impact of this term.

Here we can see quite clearly that it helps finding
collision free paths in all examples and using
all goal constraints models.
It is worth noting that the best success rates
are obtained when we use a geodesic attractor
with a strong geodesic term, confirming
the importance of modeling the motion optimization
problem with Riemannian geometry.

\section*{CONCLUSION}

We presented a methodology for modeling
motion optimization problems.
We showed that these models allow
to solve challenging path planning problem
with an interior point method.
We draw key insights from Riemannian geometry to introduce
a new objective and a constraint terms based
on the geodesic flow that lead to higher
success rates on planning problems with narrow passages.

One way to view our contribution is to see it
as a way to integrate a dynamic programming
precomputation (i.e., the geodesic flow), within
a gradient-based optimization framework.
In future work we aim to further remove this need
for solving a diffusion process and rather decompose
the workspace in primitive shapes that analytically
provide a solution to the geodesic flow such as proposed in \cite{Huber:19}.

\section*{ACKNOWLEDGMENT}
This work is partially funded by the research alliance ``System Mensch''.
The authors thank the International Max Planck Research School for Intelligent Systems (IMPRS-IS).
This research was also supported in part by National Science Foundation grants IIS-1205249, IIS-1017134, EECS0926052, the Office of Naval Research, the Okawa Foundation, and the Max-Planck-Society. Any opinions, findings,
and conclusions or recommendations expressed in this material are those of the author(s) and do not necessarily reflect
the views of the funding organizations.

\bibliographystyle{IEEEtran}
\balance
\bibliography{bibliography}

\end{document}

%% file: basic_environments.tex
\usepackage{hyperref}

\usepackage{bm}
\usepackage{amsbsy}
\usepackage{amsopn}
\usepackage{amsfonts}
\usepackage{amsmath}

\newcommand{\spt}[1]{\mathcal{#1}}

